\documentclass[sigconf]{acmart}

\AtBeginDocument{%
  }
    
\copyrightyear{2026}
\acmYear{2026}
\setcopyright{none}
\acmConference[KDD '26]{Proceedings of the 32nd ACM SIGKDD Conference on Knowledge Discovery and Data Mining V.2}{August 09--13, 2026}{Jeju Island, Republic of Korea}
\acmBooktitle{Proceedings of the 32nd ACM SIGKDD Conference on Knowledge Discovery and Data Mining V.2 (KDD '26), August 09--13, 2026, Jeju Island, Republic of Korea}
\acmDOI{10.1145/3770855.3817874}
\acmISBN{979-8-4007-2259-2/2026/08}

\usepackage{enumitem}

\begin{document}

\title[DEFINED: A Data-Efficient Computational Framework for Fine-Grained Creativity Assessment \\ in Debate Scenarios]{DEFINED: A Data-Efficient Computational Framework for Fine-Grained Creativity Assessment in Debate Scenarios}

\author{Tongzhou Yu}
\authornote{These authors contributed equally to this work.}
\email{253108120137@sii.edu.cn}

\affiliation{%
  \institution{Nanjing University}
  \city{Nanjing}
  \country{China}
}

\affiliation{%
  \institution{Shanghai Innovation Institute}
  \city{Shanghai}
  \country{China}
}

\author{Mingjia Li}
\authornotemark[1]
\email{limj@stu.ecnu.edu.cn}

\affiliation{%
  \institution{East China Normal University}
  \city{Shanghai}
  \country{China}
}

\author{Hong Qian}
\authornote{Corresponding author.}
\email{hqian@cs.ecnu.edu.cn}

\affiliation{%
  \institution{East China Normal University}
  \city{Shanghai}
  \country{China}
}

\affiliation{%
  \institution{Shanghai Innovation Institute}
  \city{Shanghai}
  \country{China}
}

\author{Wenkai Wang}
\email{52295901050@stu.ecnu.edu.cn}

\affiliation{%
  \institution{East China Normal University}
  \city{Shanghai}
  \country{China}
}

\author{Zongbao Zhang}
\email{50250936003@stu.ecnu.edu.cn}

\affiliation{%
  \institution{East China Normal University}
  \city{Shanghai}
  \country{China}
}

\affiliation{%
  \institution{Shanghai Innovation Institute}
  \city{Shanghai}
  \country{China}
}

\author{Yaoyu Jiang}
\email{10235102489@stu.ecnu.edu.cn}

\affiliation{%
  \institution{East China Normal University}
  \city{Shanghai}
  \country{China}
}

\author{Xiangfeng Wang}
\email{xfwang@cs.ecnu.edu.cn}

\affiliation{%
  \institution{East China Normal University}
  \city{Shanghai}
  \country{China}
}

\author{Aimin Zhou}
\email{amzhou@cs.ecnu.edu.cn}

\affiliation{%
  \institution{East China Normal University}
  \city{Shanghai}
  \country{China}
}
\affiliation{%
  \institution{Shanghai Innovation Institute}
  \city{Shanghai}
  \country{China}
}

\author{Jiajun Guo}
\email{jjguo@psy.ecnu.edu.cn}

\affiliation{%
  \institution{East China Normal University}
  \city{Shanghai}
  \country{China}
}

\renewcommand{\shortauthors}{Tongzhou Yu et al.}

\begin{abstract}
  Human creativity has emerged as a critical competency in the era of large language models. Assessing creativity in complex, open-ended environments is a grand challenge in data mining, currently hindered by a reliance on standardized simple tasks and the scarcity of fine-grained expert data. 
  As an ecologically valid assessment context, debate reflects multiple dimensions of creativity, encompassing both divergent thinking and convergent thinking. Moreover, debate is a data-rich domain, with a large volume of publicly accessible materials.
  Current mainstream automated scoring methods are poorly suited to complex settings such as debate, and therefore still rely on costly human evaluation.
  To this end, this paper proposes DEFINED, a data-efficient computational framework for fine-grained creativity assessment in debate scenarios. DEFINED operationalizes debate creativity through a hierarchical eight-dimensional metric system, implemented via a pre-trained autoregressive language model with a hierarchical scoring head that supports both fine-grained and coarse-grained evaluation. 
  Statements and their associated expert scores were obtained from authentic debate competitions, and a constrained data augmentation strategy was employed to address the elite bias inherent in the original data.
  DEFINED adopts a mixed-granularity training strategy enabling robust learning from limited fine-grained supervision annotated by trained graduate experts. To rigorously validate ecological validity beyond synthetic benchmarks, we incorporate an empirical study with debate-naive participants, utilizing these authentic data to serve as a qualitative case study for mid-to-low proficiency populations. Across our evaluation protocol, our scoring model achieves accurate and stable scoring, outperforming prompt-based large language model evaluators and existing debate scoring methods, while mitigating common failure modes observed in current approaches. The code for DEFINED is available on GitHub at \url{https://github.com/tzwo/DEFINED}.
\end{abstract}




\maketitle

\section{Introduction}
Creativity is widely recognized as a pivotal competence for the 21st century, serving as the engine for complex problem-solving and societal innovation~\cite{jobs2025future, oecd2024pisa, thornhill2023creativity}. The assessment of creativity constitutes a foundation for research in creativity. Consequently, the accurate and scalable assessment of creativity has become a central pursuit in psychology and education~\cite{zhang2026creative}. Traditional assessment paradigms primarily rely on Divergent Thinking (DT) tasks, such as the Alternative Uses Task (AUT)~\cite{mednick1962associative} or the Remote Associates Test (RAT)~\cite{mednick1968remote}, and self-report questionnaires, like the Creative Achievement Questionnaire~\cite{CAQ} and the Inventory of Creative Activities and Achievements ~\cite{jauk2014road}. While foundational, these methods face significant ecological validity challenges. Questionnaires are susceptible to social desirability bias and affective interference, often failing to reflect actual performance~\cite{podsakoff2003common,park2016revisiting}. Similarly, standard DT tasks, often conducted in decontextualized laboratory settings, struggle to capture the situated nature of creativity required in real-world environments~\cite{zeng2011can,yang2022creative}, where individuals must navigate complex constraints and adversarial dynamics.

To bridge the gap between psychometric assessment and real-world creative capability, debate serves as an ideal proxy for assessing ecological creativity. At its core, argumentation is not merely a linguistic exercise but a dynamic integration of creative and critical thinking~\cite{glassner2007stands}. Debate demands that participants generate novel arguments (divergent thinking) while adhering to logical coherence and rebuttal constraints (convergent thinking) under time pressure~\cite{Project_Debater,PersuasionForGood}. 
More importantly, this perspective aligns with process-analytic models of creativity, which posit that creative problem-solving involves a cycle of core processes, including problem construction, information encoding, idea generation, and idea evaluation~\cite{process_analytic_models,creative_processes}. Debate naturally instantiates this entire cycle under time pressure, rendering it a well-motivated task paradigm for the present study. 


However, transitioning from theoretical appreciation to computational assessment in debate scenarios presents three formidable challenges. First, the assessment bottleneck: manual evaluation of creativity in debate is notoriously labor-intensive and expensive. A single fine-grained annotation for a long-form debate statement averages over 40 minutes for a trained expert, rendering large-scale annotation computationally intractable~\cite{Reference-Based,SMOTE,CrEval}. Second, the ``granularity gap'' in existing resources: current computational argumentation datasets primarily focus on holistic outcomes (e.g., win/loss labels or general quality scores) rather than the specific dimensions of creativity~\cite{OpenDebateEvidence,ORCHID,IAM}. Winning a debate does not always equate to being creative; conflating these metrics obscures the cognitive mechanisms at play. Third, data distribution bias: existing corpora are dominated by professional debaters (``elite'' samples), lacking the mid-to-low proficiency examples typical of student populations or the general public~\cite{OpenDebateEvidence,ORCHID,IAM}. This domain shift severely hinders the deployment of models in educational settings.

In this work, we introduce \textbf{DEFINED}: a \textbf{d}ata-\textbf{e}fficient computational framework designed to automate \textbf{f}ine-gra\textbf{in}ed cr\textbf{e}ativity assessment in \textbf{d}ebate scenarios. We hypothesize that expert adjudicator evaluation is not a monolithic process but a composite judgment derived from distinct cognitive dimensions. The scoring model of DEFINED takes debate statements with context as input and produces both eight-dimensional fine-grained scores and a coarse-grained debate score aligned with human overall judgments. 

The contributions of the proposed DEFINED framework are threefold:
\begin{enumerate}
  \item We collect \textbf{authentic competition statements data scored by top-tier expert adjudicators of Mandarin debate}, which serve as the ground truth reflecting human cognitive judgment. To mitigate the elite bias inherent in authentic competition data and to better cover debate statements across a wide range of proficiency levels, we further adopt a \textbf{data augmentation strategy under triple constraints}, thereby enhancing the model’s generalization capability.
  \item This paper proposes an \textbf{eight-dimension metric system}, decoupling assessment into five creativity-specific dimensions (including divergent thinking and convergent thinking) and three debate-related (non-creativity) dimensions. Building on this foundation, we design a set of detailed scoring rubrics to serve as instructions to human annotators. Based on these scoring rubrics, a limited amount of fine-grained annotation data is produced. By learning the contribution of these dimensions, our scoring model attempts to reverse-engineer the cognitive process of expert adjudicators.
  \item This work introduces a \textbf{mixed-granularity training strategy} to overcome the scarcity of fine-grained labels. The extensive experiments show that by leveraging only 60 samples of fine-grained expert annotations alongside 4,000 coarse-grained samples, our scoring model achieves high-precision prediction, effectively solving the ``small data'' problem in complex psychometric modeling. We also establish a \textbf{three-modular evaluation protocol} to verify system robustness across varying proficiency levels (High vs. Mid-Low) and annotation granularities (Fine vs. Coarse). The experimental results provide comprehensive evidence that the scoring model in DEFINED predicts scores accurately and aligns with the underlying cognitive processes of human assessment.
\end{enumerate}
\begin{figure*}[ht]
\centering
\includegraphics[width=0.85\textwidth]{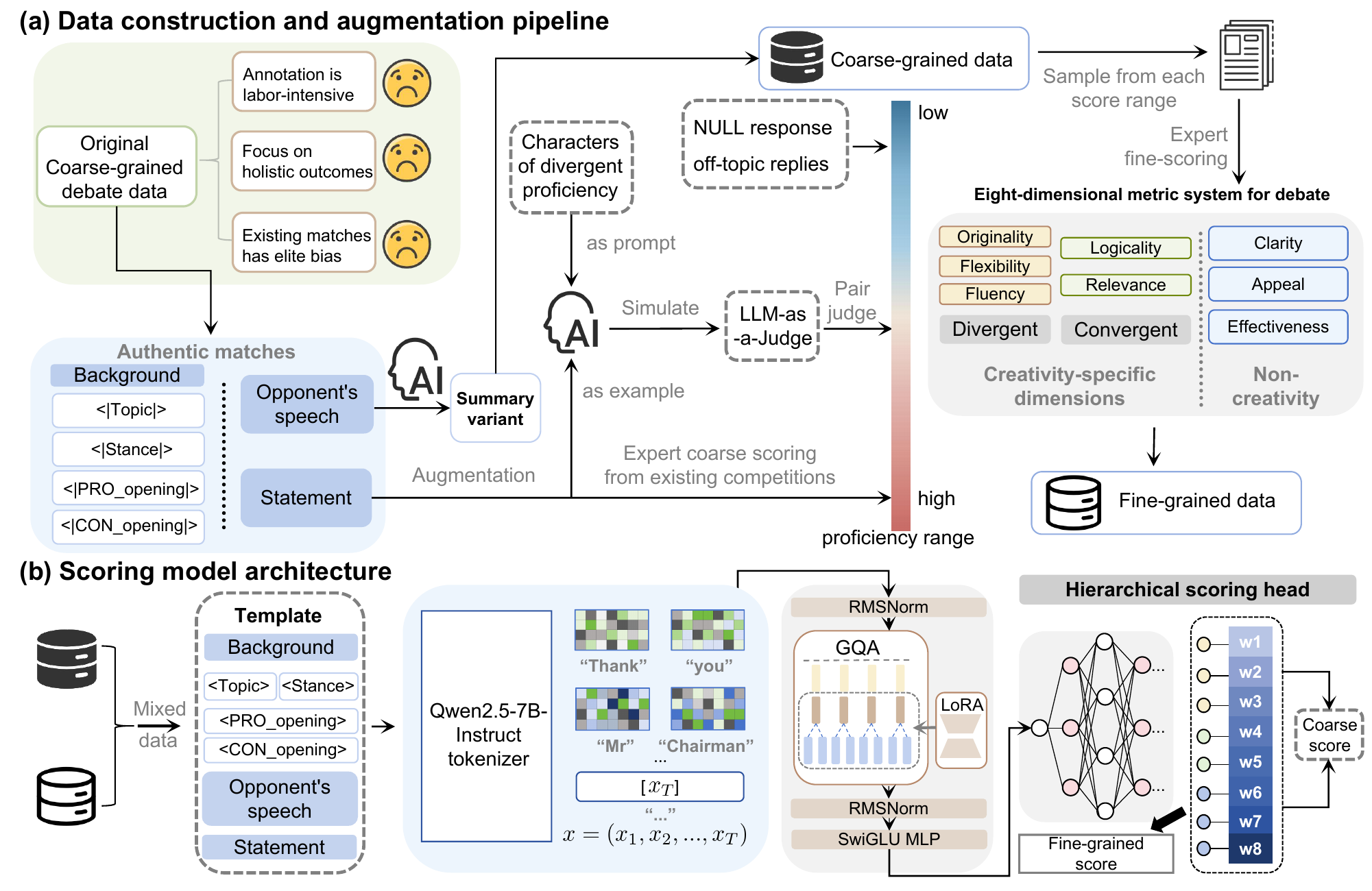}
\caption{\textbf{An overview of the proposed DEFINED framework.} 
  \textbf{(a)} Data construction and augmentation pipeline. We construct a mixed-granularity dataset by augmenting authentic high-proficiency debate competitions with synthetic samples generated under a Triple-Constraint Strategy (spanning diverse proficiency levels, noise injection, and summarization variants). A small subset is rigorously annotated by experts to provide fine-grained supervision, while the majority utilizes coarse-grained labels.
  \textbf{(b)} Scoring model architecture. The scoring model employs a pre-trained LLM (Qwen2.5-7B-Instruct) as a context-aware semantic encoder. The latent representation feeds into a specialized Hierarchical Scoring Head, which first predicts scores for eight fine-grained dimensions and subsequently aggregates them into a holistic debate score via weighted summing.}
  \Description{pic1}\label{method_pic}
\end{figure*}

\section{Related Work}
In research on automated creativity assessment, early computational approaches are predominantly grounded in semantic distance, which provides theoretical and practical foundations for assessing creative output. Rooted in associative theories of creativity, these methods operationalize originality by computing vector distances between textual concepts in high-dimensional semantic spaces, under the assumption that greater semantic distance reflects higher creative potential. Such approaches have demonstrated empirical validity in classical divergent thinking tasks including verb generation, particularly for short and decontextualized responses~\cite{Thin_slices_of_creativity,Semantic_distance}. Subsequent approaches have incorporated modern word embedding models such as Word2Vec and GloVe, which demonstrate improved performance over traditional latent semantic analysis (LSA) methods~\cite{assessment_with_SemDis}. Some recent studies have begun to emphasize the importance of context-based creativity assessment~\cite{alphacontext}. However, classical semantic models rely on static word representations and thus lack contextual sensitivity, rendering them incapable of distinguishing polysemous meanings across discourse contexts. Moreover, these methods exhibit intrinsic deficiencies when applied to long-form text: for example, LSA underestimates semantic distance as the length of response increases, a bias that runs counter to human judgments of creativity~\cite{application_by_elaboration}. As a result, semantic-distance-based measures struggle to capture the complex, multi-word semantic interactions inherent in context-dependent creative tasks~\cite{Automated_scoring}.

To overcome these limitations, fine-tuned large language models (LLMs) have emerged as one of the most human-aligned solutions for automated creativity and text quality evaluation. By attaching task-specific scoring heads to LLMs and fine-tuning them on large-scale datasets paired with human annotations, these approaches learn the latent structure of human evaluation~\cite{scoring_with_LLM}. Recent fine-tuned models such as Themis~\cite{themis} and M-Prometheus~\cite{mprometheus} further demonstrate the effectiveness of supervised learning for text assessment. Across tasks such as AUT and creative problem solving, fine-tuned LLMs have achieved correlations with human judgments exceeding 0.7, substantially outperforming semantic-distance-based approaches~\cite{luchini2025automated, Automated_scoring}. Nevertheless, the principal drawback of this paradigm lies in its heavy reliance on large volumes of high-quality human annotations. Constructing such datasets entails significant financial and temporal costs, thereby limiting scalability and hindering rapid adaptation to new creative tasks or applications in low-resource settings.

Prompt-engineering-based LLM evaluation methods have consequently been proposed as a cost-effective alternative. By embedding expert-designed scoring rubrics which explicitly define dimensions into structured prompts, these approaches enable zero-shot or few-shot creativity assessment without task-specific training data, offering clear advantages in speed and efficiency~\cite{Automated_scoring}. Empirical studies report that few-shot prompted LLMs achieve moderate correlations with human ratings (approximately 0.6) on creative problem-solving tasks, in some cases outperforming traditional semantic-distance measures while underperforming fine-tuned models~\cite{luchini2025automated}. Recent advancements in the automated evaluation of debate statement quality also adopt this method~\cite{InspireDebate, Debatrix}. However, such methods remain fundamentally constrained by intrinsic biases of LLMs (including verbosity bias, self-enhancement bias) and by instability in score calibration~\cite{LLM-as-a-Judge}. These limitations cannot be fully mitigated through prompt design alone and pose serious challenges for high-stakes applications such as talent selection or educational assessment, where fairness, objectivity, and reliability are essential.

\section{The Proposed DEFINED Framework}
\subsection{Data Collection and Preprocessing}\label{Data}
\textbf{Naturalistic Debate Corpus Construction.}
To establish a high-fidelity benchmark for ecological creativity, we collect authentic debate data from four consecutive editions (2022--2025) of \emph{Xin Guo Bian}, a premier competitive debating tournament in Mandarin debate. The tournament's adjudication standards exhibit high inter-rater reliability, serving as stable ground truth for high-quality argumentation. We operationalize the assessment task as a single-round statement evaluation: each data instance $x$ comprises contextual history $C$ (debate motion, stance, opening statements, and the opponent's preceding argument) and focal statement $r$. The corpus spans 56 distinct motions and 108 competitions, yielding 706 valid history-statement pairs obtained via high-fidelity audio transcription, with a total amount of 3,257,458 Chinese characters.

\noindent\textbf{Ground Truth Acquisition.}
For each statement, coarse-grained holistic scores are assigned by 3--5 human annotators on a 0--10 scale. To mitigate varying leniency biases among adjudicators, raw scores are standardized ($z$-score normalization) and subsequently mapped to a high-proficiency interval. This authentic, high-performing subset is denoted as $D_{\text{high}}$, representing the effective upper bound of debate performance in the target domain.

\subsection{Data Augmentation Strategy Under Triple Constraints}\label{Augmentation}
Authentic debate datasets inherently suffer from \emph{elite bias}, lacking representation of the mid-to-low proficiency spectrum typical of general educational settings. To bridge this distribution gap, we construct a synthetic dataset $D_{\text{simulate}}$ using an LLM under a proposed \emph{Triple-Constraint Augmentation Strategy}. 

\noindent\textbf{Exemplar-Based Stratified Generation.} For each target score interval $y \in [y_{\text{low}}, y_{\text{mid}}]$, we design prompt templates $p_y$ encapsulating specific competency deficiencies (e.g., logical gaps, redundancy, detailed in Appendix~\ref{Augmentation_prompt}). Conditioned on the context $C$ and referencing exemplars in $D_{\text{high}}$, the generator (Qwen2.5-72B-Instruct~\cite{Qwen2.5}) outputs candidate statements $\hat{r} = G(C, p_y)$.

\noindent\textbf{Pairwise Preference Alignment.} Prior work has shown relative ranking judgments are generally more stable and reliable than absolute scoring in LLM evaluation~\cite{Dialectical_Agent, Empirical_Analysis}. Thus, to rectify potential misalignment between generated content and target labels, we introduce a pairwise constraint. Generated samples are subjected to LLM-based pairwise comparison against anchor samples. If the inferred quality ranking $rank(\hat{r}_i, \hat{r}_j)$ contradicts the assigned pseudo-labels, score swaps are applied to ensure label consistency. In addition, we conduct a small-scale human validation: expert annotators inspected 10 groups of synthetic samples, and the corrected rankings are found to align well with human judgments.

\noindent\textbf{Noise Injection and Contextual Variation.} We further enrich the data by introducing negative samples $D_{\text{neg}}$ (e.g., null responses) and a summarization variant $D_{\text{summary}}$, where the opponent's statement is logically compressed by Qwen2.5-7B-Instruct to force the model to attend to core arguments. The final training set $D_{\text{train}}$ combines $D_{\text{high}}$, $D_{\text{simulate}}$, $D_{\text{neg}}$, and $D_{\text{summary}}$, followed by a self-correction pass where a preliminarily trained model re-scores a subset of noisy samples to improve label robustness~\cite{Robust}.

\subsection{Eight-Dimensional Metric System}\label{Scheme}
\textbf{Psychometric System Definition.}
We synthesize a hierarchical evaluation protocol bridging the process-analytic model of creative problem-solving~\cite{ process_analytic_models,creative_processes} and computational debate metrics~\cite{ClaimRev, 15-Dimensional}. The system decouples assessment into five \emph{Creativity-Specific Dimensions}: divergent thinking (\textbf{Fluency}, \textbf{Originality}, \textbf{Flexibility}) and convergent thinking (\textbf{Logicality}, \textbf{Relevance}), and three \emph{Non-Creativity Dimensions}: \textbf{Effectiveness}, \textbf{Clarity}, \textbf{Appeal}. We provide the detailed definition in Appendix~\ref{definition}.


Based on this metric system, we curate a fine-grained dataset $D_{\text{fine}}$ ($N=120$) balanced across $D_{\text{high}}$ and $D_{\text{simulate}}$, annotated by ten trained graduate experts (Cronbach's alpha = .90-.98). This subset provides the supervision signal for the multi-dimensional score.

\begin{figure*}[ht]
\centering
\includegraphics[width=0.9\textwidth]{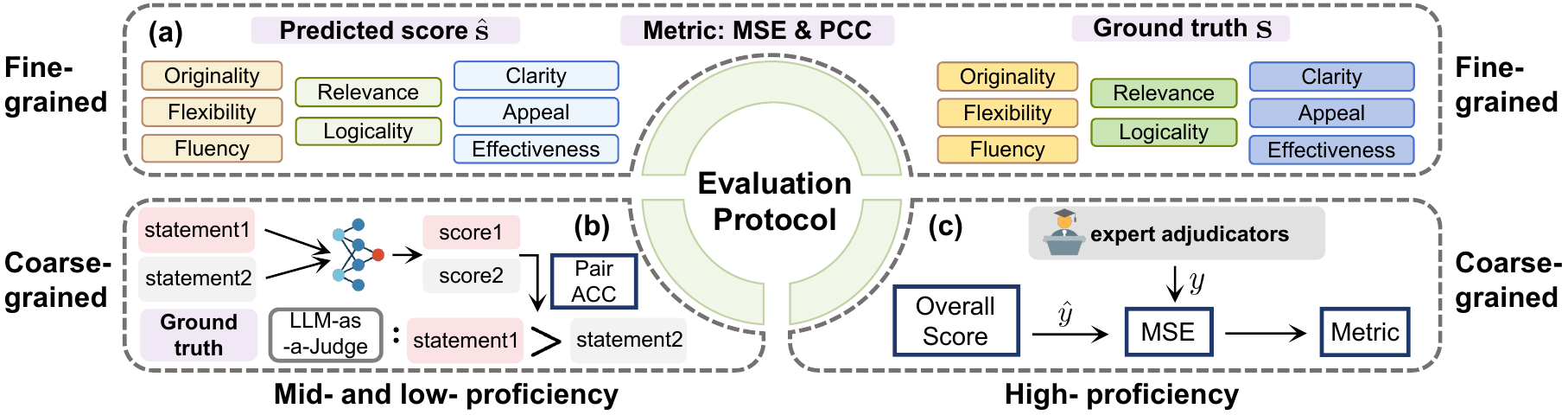}
\caption{\textbf{The three-modular evaluation protocol to assess the performance of scoring model.} 
  This evaluation protocol validates system robustness across varying annotation granularities (Fine vs. Coarse) and proficiency levels (High vs. Mid-Low).
  \textbf{(a)} Fine-grained evaluation measures the alignment of eight-dimensional predictions with human expert annotations using Mean Squared Error (MSE) and Pearson Correlation Coefficient (PCC).
  \textbf{(b)} Coarse-grained evaluation in the mid-to-low proficiency range. Given the instability of absolute scoring for lower-quality responses, we utilize Pairwise Accuracy (ACC) to assess the model's capability to correctly rank synthetic sample pairs.
  \textbf{(c)} Coarse-grained evaluation in the high-proficiency range. We benchmark the model's predictions against authentic top-tier expert adjudicator scores using MSE.}
  \Description{pic}
\label{eval_pic}
\end{figure*}
\subsection{The Architecture of Scoring Model in DEFINED}\label{architecture}
\textbf{Overview.}
The scoring model of DEFINED is designed to give both fine-grained dimensional scores and a holistic quality score for debate. The architecture consists of a semantic encoder based on a pre-trained autoregressive language model and a hierarchical scoring head.


\noindent\textbf{Context-Aware Semantic Encoder.}
In DEFINED, the autoregressive language model serves as the semantic encoder, responsible for mapping long-form inputs in debate into high-dimensional hidden representations. We adopt Qwen2.5‑7B‑Instruct as backbone model.

The model input is constructed by concatenating background information and the current-round statement according to a fixed template, yielding a single sequence of text. This design enables the model to explicitly exploit the full debate context when encoding the current statement, thereby capturing dependencies among arguments and their adversarial dynamics. 

The concatenated text is tokenized using the Qwen2.5‑7B‑Instruct tokenizer at the subword level and converted into a token sequence $x = (x_1, x_2, \dots, x_T)$,
where $T$ denotes the sequence length. 
The input token sequence is first mapped into a continuous vector space through an embedding layer, producing the representation $h_0 \in \mathbb{R}^{T \times d}$, 
where $d$ is the hidden dimensionality of the model. 
The model consists of a stack of Transformer decoder layers. The hidden state update at the $l$-th layer can be formalized as:
\begin{equation}
h_l = \mathrm{TransformerLayer}^{(l)}(h_{l-1})\,,
\end{equation}
where each layer comprises a multi-head self-attention mechanism and a feed-forward network (FFN), combined with residual connections and layer normalization. The self-attention mechanism allows the model to perform weighted aggregation over tokens at different positions in the sequence, thereby modeling argumentative structure, semantic coherence, and contextual dependencies.
After the final Transformer decoder layer, the model outputs the sequence of hidden states:
\begin{equation}
H = (h_{1}, h_{2}, \dots, h_{T}) \in \mathbb{R}^{T \times d}\,.
\end{equation}
In this work, we take the hidden state of the final token $h_{last} = H[T, :]$ in the sequence as a global representation, which serves as a high-level semantic feature for subsequent modules.

\noindent\textbf{Hierarchical Scoring Head.}
The custom scoring head is responsible for mapping the high-dimensional semantic representations into multi-dimensional creativity assessment, and for further deriving a coarse-grained overall score that is consistent with existing debate annotation schemes. This module adopts a multi-layer neural network architecture and explicitly models the hierarchical relationship between fine-grained and coarse-grained scores, thereby accommodating supervision signals of mixed granularity.

The input to this module is the high-level semantic representation $h_{\text{last}}$ obtained from the autoregressive language model. This representation encodes the overall semantics, argumentative structure, and contextual dependencies of the current utterance under the given debate setting. The custom scoring head consists of two sequential components:  
(1) a multi-layer non-linear mapping network for predicting fine-grained scores across eight evaluation dimensions; and  
(2) an explainable aggregator that combines the eight-dimensional scores into a single scalar debate score.

First, $h_{\text{last}}$ is transformed through a stack of FFN layers. The network is composed of multiple fully connected layers, each followed by a SiLU (Sigmoid Linear Unit) activation function:
\begin{equation}
h^{(k)} = \mathrm{SiLU}\!\left(W^{(k)} h^{(k-1)}\right)\,,
\end{equation}
where $h^{(0)} = h_{\text{last}}$, and $W^{(k)}$ denotes the weight matrix of the $k$-th layer. At the final layer, the hidden representation is projected into an eight-dimensional vector:
\begin{equation}
\hat{\mathbf{s}} = (\hat{s}_1, \hat{s}_2, \dots, \hat{s}_8) \in \mathbb{R}^{8}\,,
\end{equation}
where each component corresponds to the predicted score of one dimension in the debate metric system. This design enables the model to learn differentiated responses across dimensions while sharing a common semantic representation.

To produce an overall score aligned with overall debate adjudication, the eight fine-grained scores are further passed to an aggregator module. The aggregator computes a weighted sum of the dimension-wise scores, which can be formalized as $\hat{y} = \sum_{i=1}^{8} w_i \hat{s}_i$,
where $\mathbf{w} = (w_1, \dots, w_8)$ are learnable aggregation weights. This structure is intended to emulate the weighting and integration of multiple evaluation dimensions performed by expert adjudicators when assigning an overall score.
By introducing this aggregation layer, the model is able to output both fine-grained eight-dimensional scores and a coarse-grained overall score within a single forward pass, thereby naturally aligning with datasets annotated at different levels of granularity.

\noindent\textbf{Hybrid Loss Function.}
To leverage mixed-granularity supervision, we optimize a hybrid objective function. For samples in $D_{\text{coarse}}$ (and augmented subsets), we minimize the Mean Squared Error (MSE) on the holistic score. For samples in $D_{\text{fine}}$, we additionally minimize the sum of MSE across all dimensions. The total loss is defined as:
\begin{equation}
    \mathcal{L} = \mathbb{I}_{y \in D}\text{MSE}(\hat{y}, y) + \lambda \cdot \mathbb{I}_{s \in D_{\text{fine}}}\sum_{j=1}^{8} \text{MSE}(\hat{s}_j, s_j)\,,
\end{equation}
where $\mathbb{I}$ is the indicator function and $\lambda$ balances the multi-task learning.


During training, the appropriate loss term is dynamically selected according to the annotation granularity of each sample and used for back-propagation. This design allows the model to jointly leverage large-scale, easily obtainable coarse-grained data and a limited amount of fine-grained data within a shared parameter space, enabling data-efficient learning of multidimensional scoring.
\begin{table*}[t]
\centering
\caption{Pearson correlation coefficients (PCC) between model predictions and expert scores for each dimension. Higher values indicate stronger agreement with human evaluators.}
\label{fine_PCC}
\setlength{\tabcolsep}{4pt}
\begin{tabular}{l|cccccccc|c}
\toprule
\textbf{Model} & \textbf{Fluency} & \textbf{Originality} & \textbf{Flexibility} & \textbf{Logicality} & \textbf{Relevance} & \textbf{Effectiveness} & \textbf{Clarity} & \textbf{Appeal} & \textbf{Average}\\
\midrule
Gemini-2.5-pro & 0.71 & 0.77 & 0.76 & 0.71 & 0.67 & 0.71 & 0.44 & 0.61 & 0.67\\
GPT-4o         & 0.70 & 0.64 & 0.67 & 0.68 & 0.70 & 0.70 & 0.58 & 0.71 & 0.67\\
Qwen3-max-preview  & 0.74 & 0.77 & 0.75 & 0.69 & 0.75 & 0.75 & 0.50 & 0.74 & 0.71\\
Deepseek-R1    & 0.79 & 0.77 & 0.74 & 0.74 & 0.77 & 0.83 & 0.55 & 0.76 & 0.74\\
\midrule
M-Prometheus-7B    & 0.59 & 0.59 & 0.53 & 0.60 & 0.57 & 0.62 & 0.56 & 0.55 & 0.58\\
Themis    & -0.02 & 0.31 & 0.20 & 0.07 & 0.16 & 0.37 & -0.22 & 0.14 & 0.12\\
\midrule
\textbf{DEFINED} & \textbf{0.96} & \textbf{0.96} & \textbf{0.94} & \textbf{0.96} & \textbf{0.96} & \textbf{0.96} & \textbf{0.93} & \textbf{0.97} & \textbf{0.96}\\
\bottomrule
\end{tabular}
\end{table*}
\begin{table*}[t]
\centering
\caption{Mean squared error (MSE) of fine-grained score predictions for DEFINED and four representative LLM-based evaluators. Lower MSE indicates closer numerical alignment with expert judgments.}
\label{fine_MSE}
\setlength{\tabcolsep}{4pt}
\begin{tabular}{l|cccccccc|c}
\toprule
\textbf{Model} & \textbf{Fluency} & \textbf{Originality} & \textbf{Flexibility} & \textbf{Logicality} & \textbf{Relevance} & \textbf{Effectiveness} & \textbf{Clarity} & \textbf{Appeal} & \textbf{Average}\\
\midrule
Gemini-2.5-pro & 358.75 & 465.68 & 444.08 & 419.44 & 506.66 & 555.39 & 597.76 & 550.12 & 487.24\\
GPT-4o         & 305.51 & 347.46 & 400.42 & 372.10 & 355.73 & 369.64 & 395.76 & 332.93 & 359.94\\
Qwen3-max-preview  & 243.64 & 208.93 & 251.24 & 300.58 & 231.15 & 230.19 & 440.25 & 276.44 & 272.80\\
Deepseek-R1    & 298.73 & 207.10 & 226.75 & 240.95 & 225.88 & 166.90 & 332.37 & 271.61 & 246.29\\
\midrule
M-Prometheus-7B    & 352.50 & 402.72 & 465.48 & 405.02 & 488.47 & 416.18 & 407.10 & 510.42 & 430.99\\
Themis    & 1943.67 & 1143.48 & 1148.60 & 1830.02 & 1576.83 & 963.40 & 2366.62 & 1483.35 & 1557.00\\
\midrule
\textbf{DEFINED}   & \textbf{42.24}  & \textbf{34.04}  & \textbf{59.60}  & \textbf{46.72}  & \textbf{35.98}  & \textbf{35.76}  & \textbf{57.01}  & \textbf{33.36} & \textbf{43.09}\\
\bottomrule
\end{tabular}
\end{table*}

\subsection{Evaluation Protocol}\label{Evaluation}
We implement a rigorous \emph{Three-Modular Evaluation Protocol} to assess the performance across proficiency levels (High vs. Mid-Low) and granularities (Fine vs. Coarse).

For coarse-grained evaluation, we reserve 10\% of the authentic competition data ($D_{\text{high}}$) and the corresponding summary variants as validation data. We benchmark DEFINED against state-of-the-art approaches of debate evaluation (e.g., InspireDebate~\cite{InspireDebate}, Debatrix~\cite{Debatrix}). Performance is quantified using MSE as metric of high-score range data and Pairwise Accuracy for mid- and low-score range data, leveraging the high reliability of comparative labels in the synthetic subset $D_{\text{simulate}}$.

For fine-grained evaluation, $D_{\text{fine}}$ is split (50/50) into training and validation sets, ensuring no overlap with the coarse-grained training data. We compare our scoring model with LLM evaluators (e.g., GPT-4o) using well-designed prompt templates derived from the expert-defined evaluation protocol. These prompts include detailed scoring rubrics, operational definitions, scoring checkpoints, and representative high-/low-score examples to improve scoring consistency and instruction following. The complete prompt template is provided in Appendix~\ref{fine_prompt}. We adopt MSE and Pearson correlation coefficient (PCC) to evaluate prediction accuracy and alignment with the cognitive dimensions of human annotators.

\section{Experiment}
In this section, we present the experimental results under the proposed three-modular evaluation protocol (Sec.~\ref{Score_fine}-\ref{high_range}), case study of typical bias patterns in LLMs (Sec.~\ref{Case_stuty}), ablation study that isolates the coarse-grained data (Sec.~\ref{Ablation}) and interpretability analysis of dimension weights (Sec.~\ref{Interpretability}). Given that each evaluation setting involves different data sources and performance metrics, we provide a detailed description for each part. The code for DEFINED is available on GitHub at \url{https://github.com/tzwo/DEFINED}.
\subsection{Experimental Setup and Costs}
All experiments are conducted by fine-tuning Qwen2.5-7B-Instruct as the semantic backbone of DEFINED. Training is performed on 8 NVIDIA H100 GPUs. The total training time is 6{,}453 seconds, reflecting the data-efficient nature of the proposed framework. The main training hyperparameters are set as follows: learning rate $4\times10^{-4}$, maximum sequence length 8{,}192, training batch size 2 with gradient accumulation over 8 steps, validation batch size 8, and a total of 30 training epochs. Model evaluation is performed every 30 steps to monitor convergence throughout training.

To enable parameter-efficient adaptation under limited fine-grained supervision, we adopt Low-Rank Adaptation (LoRA)~\cite{LoRA} for model fine-tuning. Specifically, the LoRA rank is set to 8 with a scaling factor (alpha) of 32 and a dropout rate of 0.1. LoRA modules are injected into attention-related projection layers, including \texttt{q\_proj} and \texttt{v\_proj}, which play a central role in contextual representation learning. Optimization is carried out using the AdamW optimizer with $\beta_1=0.9$, $\beta_2=0.999$, and $\epsilon=1\times10^{-8}$. 

At inference time, \textsc{DEFINED} demonstrates low latency and favorable scalability. Evaluation on the test set is conducted using four NVIDIA H100 GPUs with a batch size of 32. Under this configuration, the model achieves an average inference time of 0.045 seconds per sample, enabling efficient large-scale deployment. Taken together, these results indicate that \textsc{DEFINED} attains strong performance while maintaining moderate training cost and fast inference, making it practical for real-world educational and assessment scenarios where both accuracy and efficiency are critical.
\begin{figure*}[ht]
\centering
\includegraphics[width=0.9\textwidth]{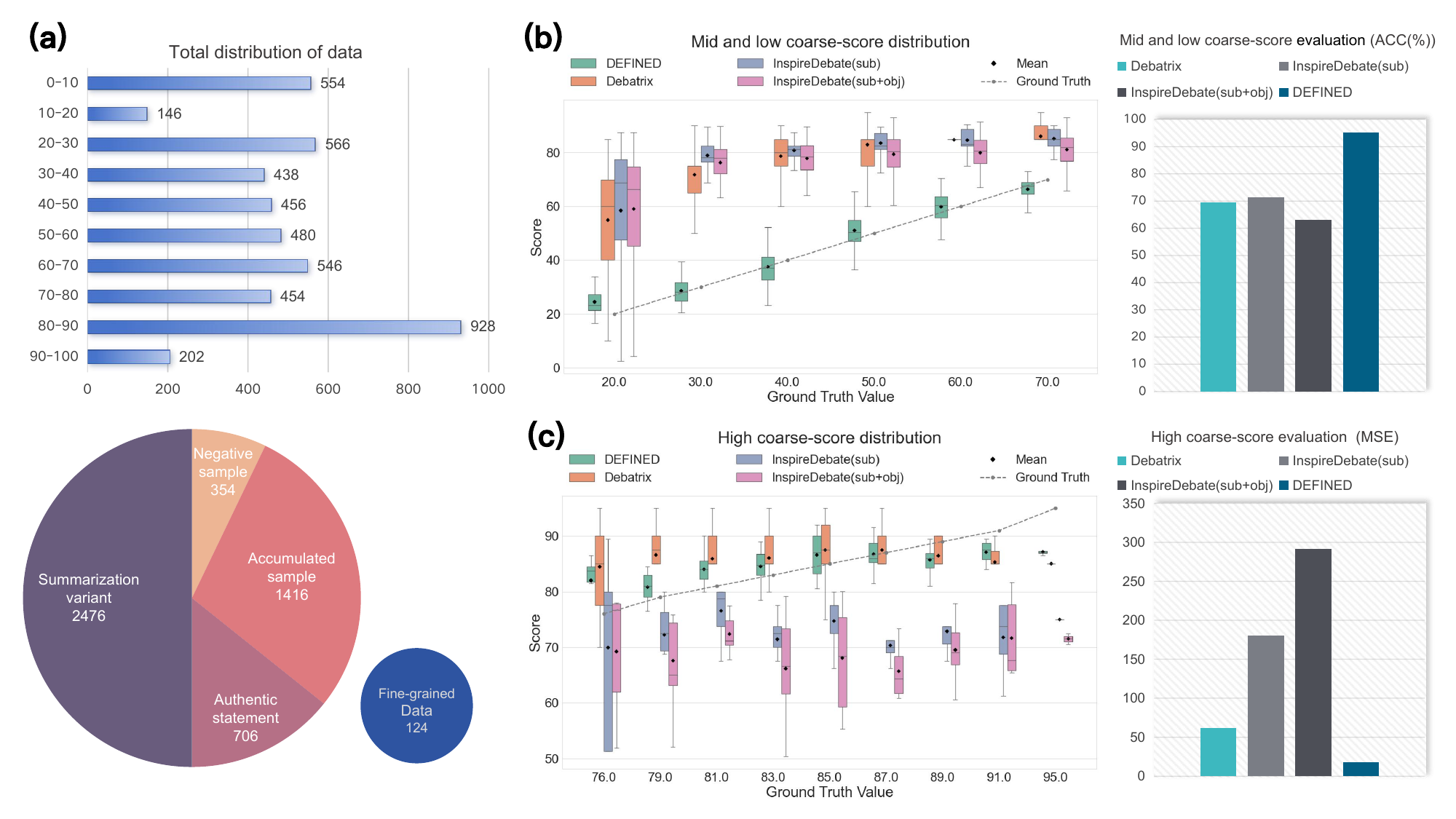}
\caption{\textbf{Performance of coarse-grained debate assessment across proficiency spectrums.} 
  \textbf{(a)} Dataset composition. The distribution of authentic high-proficiency samples alongside synthetic mid-to-low proficiency variants generated via our augmentation strategy.
  \textbf{(b)} Evaluation on mid-to-low proficiency synthetic data. Pairwise agreement ACC shows the proposed framework's robustness in distinguishing relative quality among weaker responses, effectively mitigating the scoring instability observed in baseline models.
  \textbf{(c)} Evaluation on high-proficiency authentic competition data. DEFINED achieves significantly lower MSE compared to LLM-based baselines (Debatrix, InspireDebate), indicating superior alignment with expert adjudicators.}
  \Description{pic}
\label{coarse_result}
\end{figure*}
\subsection{Accurate Predictions of Scoring Model in Fine-Grained Score}\label{Score_fine}
We first evaluate DEFINED's core contribution, i.e., multi-dimensional creativity assessment based on the fine-grained evaluation protocol (Fig.~\ref{eval_pic}(a)). We compare DEFINED against two categories of baselines: (1) foundation model, including Gemini-2.5-Pro, GPT-4o, Qwen3-Max, and DeepSeek-R1, which rely on prompt-based scoring; and (2) fine-tuned models for text assessment, including Themis and M-Prometheus. All models are evaluated on the same validation dataset using the expert-defined scoring rubrics as prompts. We adopt multiple metrics, including MSE and PCC, to jointly measure prediction accuracy and alignment with human annotators.

As shown in Tab.~\ref{fine_PCC}, DEFINED consistently outperforms all baseline evaluators across all eight dimensions. While API-based models exhibit moderate correlations with expert scores (average PCC ranging from 0.67 to 0.74), fine-tuned models show weaker alignment in this setting. Specifically, M-Prometheus-7B achieves only moderate agreement (average PCC 0.58), while Themis performs poorly overall (average PCC 0.12). 
In contrast, DEFINED achieves substantially stronger alignment, with an average PCC of 0.96, indicating near-expert-level internal consistency. 

Correspondingly, DEFINED reduces the average MSE by an order of magnitude compared with the best-performing API baseline (mean MSE 43.09 versus 246.29 for DeepSeek-R1), and dramatically outperforms evaluator-oriented models such as M-Prometheus-7B (430.99) and Themis (1557.00), demonstrating significantly improved numerical accuracy in fine-grained scoring (Tab.~\ref{fine_MSE}).

Among the API-based baselines, Qwen and DeepSeek outperform Gemini and GPT across most dimensions. This gap is possibly attributable to the stronger Chinese language understanding and discourse modeling capabilities of Qwen and DeepSeek, which are trained with a heavier emphasis on high-quality Chinese corpora.

These results demonstrate DEFINED not only achieves high predictive accuracy at the aggregate level but also provides stable, and human-aligned fine-grained creativity scores across dimensions. DEFINED has advantages over mainstream open-source and closed-source models in the task of fine-grained scoring of creativity. 

For the 10\% coarse-grained setting, we performed two additional re-samplings, yielding MSE values of 44.43 ± 1.40 across runs. For the fine-grained setting, we conducted one additional re-sampling, resulting in an MSE of 42.35 ± 0.92. In contrast, API-based methods still exhibit substantially higher errors (MSE > 350). These results demonstrate only minor variation across runs, while maintaining consistent performance advantages over baselines.
\begin{table*}[t]
\centering
\caption{Ablation study on mixed-granularity supervision using mean squared error (MSE).}
\label{ablation_mse}
\resizebox{\textwidth}{!}{
\begin{tabular}{l|cccccccc|c}
\toprule
\textbf{Method} 
& \textbf{Logicality} 
& \textbf{Appeal} 
& \textbf{Relevance} 
& \textbf{Clarity} 
& \textbf{Effectiveness} 
& \textbf{Flexibility} 
& \textbf{Originality} 
& \textbf{Fluency} 
& \textbf{Average} \\
\midrule
Fine-grained Only 
& 140.35 
& 109.66 
& 125.33 
& 97.31 
& 97.87 
& 102.41 
& 154.66 
& 82.76 
& 113.79 \\
w/o Simulation
& 563.49
& 465.08
& 556.72
& 462.33 
& 516.56 
& 544.17
& 540.05
& 476.85
& 515.66 \\
w/o Summarization
& 81.48
& 100.82
& 90.17 
& 81.46 
& 64.37 
& 104.26
& 94.78
& 114.10
& 91.43 \\
\midrule
DEFINED (Mixed) 
& \textbf{46.72} 
& \textbf{33.36} 
& \textbf{35.98} 
& \textbf{57.01} 
& \textbf{35.76} 
& \textbf{59.60} 
& \textbf{34.04} 
& \textbf{42.24} 
& \textbf{43.09} \\
\bottomrule
\end{tabular}
}
\end{table*}
\subsection{Accurate Coarse-Grained Pairwise Comparisons in Mid- and Low-Score Range}\label{low_range}

As illustrated in the mid-to-low proficiency module (Fig.~\ref{eval_pic}(b)), absolute ground truth scores are unstable for lower-quality synthetic responses. Therefore, we adopt Pairwise Agreement ACC as the primary metric. For each competition in the set (Fig.~\ref{coarse_result}(a)), we generate an additional set of samples for validation. The original pairwise ordering is used as a ground truth. Model performance is therefore evaluated using Pairwise Agreement ACC, defined as the proportion of statement pairs for which the predicted score ordering matches the original pairwise preference (Fig.~\ref{eval_pic}(b)). We compare DEFINED with two representative debate assessment methods, Debatrix and InspireDebate, both of which rely on LLM as judges.

The compared methods differ in their dimensional formulations, which are used as prompts to call LLM to get scores. InspireDebate decomposes debate quality into subjective dimensions (including Emotional Appeal, Argument Clarity, Argument Arrangement, and Topic Relevance) and objective dimensions such as Fact Authenticity and Logical Validity, which are then aggregated by means of averaging. In contrast, Debatrix organizes evaluation signals into broader categories of Argument, Source, and Language. 

As shown in Fig.~\ref{coarse_result}(b), DEFINED achieves a pairwise accuracy of 95.2\% in the mid- and low-score range, outperforming baseline methods. In contrast, Debatrix using GPT-4o achieves an accuracy of 69.4\%, while InspireDebate reaches 71.4\% when using only subjective dimensions and 63.2\% when combining subjective and objective dimensions. The margin observe between DEFINED and the baselines indicates that DEFINED more reliably distinguishes relative quality differences among mid- and low-performing statements.

Although both Debatrix and InspireDebate exhibit moderately acceptable performance in relative ordering, their absolute scores remain highly unstable in this range (box plots of Fig.~\ref{coarse_result}). In particular, even when the evaluated statement is of clearly inferior quality, API-based methods frequently assign inflated absolute scores (for example, scores approaching 85 or 90), while some high-quality statements in authentic competitions receive scores around 75. The upward bias in absolute scoring substantially limits the practical utility of these methods in educational or formative assessment scenarios, where distinguishing mediocre from strong performance is critical. In the mid- and low-score range, naive LLM-based scoring pipelines tend to overestimate performance and blur meaningful distinctions. By contrast, DEFINED demonstrates strong consistency in both relative ranking and absolute score, reflecting its ability to anchor coarse-grained predictions to expert adjudicator distributions learned from authentic competition data.

\subsection{Accurate Coarse-Grained Debate Scoring in High-Score Range}\label{high_range}
Following high-proficiency evaluation module defined in Fig.~\ref{eval_pic}(c), we assess the model's ability to approximate expert adjudicators in authentic debate scenarios. We randomly select 10\% of the competitions from the dataset as a validation set and focus on debates drawn from authentic competition data (Fig.~\ref{coarse_result}(a)). All methods are evaluated under identical data, and mean squared error (MSE) with respect to expert scores is used as the evaluation metric (Fig.~\ref{eval_pic}(c)).

As shown in Fig.~\ref{coarse_result}(c), DEFINED outperforms all baseline methods in the high-score range, achieving an MSE of 18.23, compared with 61.94 for Debatrix (GPT-4o) and 180.53 and 291.16 for InspireDebate using subjective-only and combined subjective–objective dimensions, respectively (Fig.~\ref{coarse_result}(c)). The reduction is over 70\% relative to Debatrix and over 90\% relative to InspireDebate in MSE. This performance gap indicates that DEFINED achieves a closer numerical alignment with expert adjudicator judgments in authentic competition data. Moreover, DEFINED contains only 7B parameters, which is smaller than the mainstream models used for comparison, requiring fewer computational resources during inference.

\subsection{Case Study of Typical Bias Patterns in the LLM’s Scoring Behavior}\label{Case_stuty}
To further elucidate the behavioral differences between existing LLM-based evaluators and our proposed DEFINED framework, we conduct a qualitative case study on real debate utterances. We construct a specific experimental scenario and collect speeches from 60 debate-naive participants (32 females, 28 males, undergraduate or graduate students) under this setting. These responses are evaluated using both DEFINED and representative LLMs. By examining the model‑generated scores alongside the fine‑grained results in the validation set, we manually inspect the three samples exhibiting the largest discrepancies across evaluation dimensions (detailed in Appendix~\ref{case_study}). This analysis allows us to identify several representative cases that illustrate typical bias patterns in the LLM’s scoring behavior: score instability, surface-level pattern matching, and absolute score mismatch, while highlighting the robustness and interpretability of DEFINED.

First, score instability under contextual perturbation emerges as a critical limitation of existing LLM-based evaluators. As illustrated in our examples, identical debate speeches evaluated under slightly altered opponent's speech can lead to score deviations exceeding 30 points on the same dimension for several strong baseline models. Such volatility suggests that these models do not focus on the intrinsic quality of discourse while distracted by background information. In contrast, DEFINED produces highly consistent scores across these perturbations. This stability is particularly crucial in realistic evaluation scenarios to score the target statement.

Second, we observe pattern-matching in baseline models. Statements that explicitly adopt formal discourse markers such as ``at first'', ``second'', ``finally'' are frequently rewarded with inflated fluency scores, even when their underlying arguments are shallow or repetitive. This suggests that many models conflate rhetorical templates with fluency. DEFINED can understand the deep argumentative structure of statements, rather than relying on superficial structural words, assigning more reasonable scores accordingly. 

Third, existing methods exhibit distortions in absolute score, leading to poor discrimination between low-quality and high-quality speeches. In several cases, texts with evident issues, such as oversimplified reasoning or lack of engagement with the opposing argument, receive scores comparable to, or even higher than, rhetorically complex and substantively rich speeches. This undermines the evaluative utility of automated systems in competitive or high-stakes settings. DEFINED produces well-separated score distributions: high-quality debate speeches are consistently assigned higher scores across relevant dimensions, while lower-quality texts are clearly distinguished. This suggests that DEFINED learns a more human-aligned internal scale of creative and argumentative quality.

Taken together, these cases demonstrate that DEFINED not only improves correlation with human judgments, but also addresses deeper structural weaknesses of existing LLM-based evaluators, offering a more reliable and principled solution for fine-grained creativity assessment in debate contexts.

\subsection{Ablation Study}\label{Ablation}
To validate the design choice of mixed coarse–fine-grained supervision in DEFINED, we conduct an ablation study that isolates the contribution of coarse-grained training signals. In this ablation setting, the model is trained exclusively on a limited set of fine-grained annotations. All other components of the architecture and optimization procedure are kept identical to the full model.

Tab.~\ref{ablation_mse} reports the dimension-wise MSE of the full DEFINED model compared with the ablated variant trained using only fine-grained creativity scores. Across all eight dimensions, removing coarse-grained supervision leads to a substantial degradation in numerical accuracy. The average MSE increases from 43.09 in the full DEFINED model to 113.79 under fine-grained-only training, corresponding to a relative error increase of more than 2.6$\times$.

We further analyze the role of the data augmentation strategy. Removing simulation data causes severe degradation in the mid- and low-score range: the model frequently fails to assign reasonable scores to lower-quality responses, with prediction deviations often exceeding 30 points. This finding confirms that simulated debate data is essential for mitigating elite bias introduced by expert-only annotations. In addition, removing the summarization variant increases the average MSE to approximately 90, indicating that the model is highly sensitive to contextual diversity. These results collectively demonstrate that mixed-granularity supervision is critical for robust and accurate creativity assessment. Since coarse-grained data is substantially easier to obtain and scale, it effectively improves both the generalization ability and numerical stability of DEFINED.
\subsection{Interpretability Analysis}\label{Interpretability}
The learned dimension weights---Effectiveness (0.1826), Fluency (0.1689), Originality (0.1396), Clarity (0.1221), Appeal (0.1016), Flexibility (0.0977), Relevance (0.0928), and Logicality (0.0923)---reveal a clear alignment with human adjudication in debate. Effectiveness receives the highest weight, indicating that human evaluators adopt a fundamentally outcome-oriented judgment strategy: arguments are primarily valued by whether they meaningfully advance the debate or successfully address the opponent’s core claims. Fluency and Originality are also assigned relatively high weights, reflecting that evaluators’ first cognitive response is whether a speech ``has substance''---that is, whether it generates multiple non-redundant ideas rather than merely repeating previously stated content.

By contrast, Logicality and Relevance receive lower weights, not because they are unimportant, but they function as threshold dimensions in human cognition. Logical failure leads to immediate penalization, whereas logical adequacy does not necessarily yield additional credit. Notably, Clarity is weighted higher than Logicality, suggesting that comprehensibility takes precedence over formal reasoning rigor: arguments that are easier to understand are favored over those that are logically sound but difficult to follow. Overall, this weighting structure indicates that \textsc{DEFINED} internalizes a human-like evaluation process that prioritizes practical impact and communicative effectiveness over purely formal correctness.

\section{Conclusion \& Discussion}
In this work, we proposed DEFINED, a data-efficient framework that operationalizes fine-grained creativity assessment within complex debate scenarios. By decomposing holistic expert judgments into a hierarchical eight-dimensional metric system, our model effectively reverse-engineers the cognitive process of evaluation, distinguishing specific creativity dimensions from general quality factors. Crucially, the integration of a mixed-granularity training strategy and constrained data augmentation addresses the persistent challenges of annotation scarcity and elite-data bias in computational social science. This architectural design enables the model to leverage abundant coarse-grained signals alongside limited fine-grained supervision, achieving robust, scalable, and ecologically valid scoring that significantly outperforms existing methods.

Despite these advancements, certain methodological constraints and future directions warrant discussion. First, owing to the unique mixed-granularity supervision paradigm proposed here, conventional supervised fine-tuning methods cannot be directly adapted as baselines. Instead, we ensured rigorous comparative analysis against state-of-the-art API-based LLM evaluators, utilizing expert-verified prompts strictly aligned with human annotation guidelines to guarantee fair benchmarking. Challenges also remain regarding cross-domain generalization and potential stylistic artifacts in synthetic data. Future research should focus on identifying task-invariant creativity features to enable transfer learning across disparate open-ended tasks. Beyond debate, the proposed framework offers a paradigmatic strategy for mining complex psychological constructs under data-scarce conditions: by anchoring fine-grained indicators to large-scale coarse signals, it bridges the gap between psychometric theory and scalable data mining applications.

\section{Acknowledgments}
This work is supported by the Shanghai Municipal Special Program for Basic Research on General AI Foundation Models (Grant No. 2025SHZDZX026D08) and Fundamental Research Funds for the Central Universities (Grant No. 2026ECNU-WLJC009).
\bibliographystyle{ACM-Reference-Format}
\bibliography{sample-base}

@inproceedings{jobs2025future,
  title={Future of jobs report 2025},
  author={Leopold, T and Di Battista, Attilio and Jativa, Ximena and Sharma, Shuvasish and Li, R and Grayling, S},
  booktitle={World Economic Forum},
  year={2025},
  address={Geneva, Switzerland}
}

@book{oecd2024pisa,
  title={PISA 2022 Results (Volume III): Creative minds, creative schools},
  author={OECD},
  year={2024},
  publisher={OECD Publications Centre},
  address={Paris, France}
}

@article{thornhill2023creativity,
  title={Creativity, critical thinking, communication, and collaboration: Assessment, certification, and promotion of 21st century skills for the future of work and education},
  author={Thornhill-Miller, Branden and Camarda, Ana{\"e}lle and Mercier, Maxence and Burkhardt, Jean-Marie and Morisseau, Tiffany and Bourgeois-Bougrine, Samira and Vinchon, Florent and El Hayek, Stephanie and Augereau-Landais, Myriam and Mourey, Florence and others},
  journal={Journal of Intelligence},
  volume={11},
  number={3},
  pages={54},
  year={2023}
}

@article{mednick1962associative,
  title={The associative basis of the creative process},
  author={Mednick, Sarnoff},
  journal={Psychological review},
  volume={69},
  number={3},
  pages={220},
  year={1962}
}

@article{mednick1968remote,
  title={Remote associates test},
  author={Mednick, Martha T and Halpern, Sharon},
  journal={Psychological Review},
  year={1968}
}

@article{CAQ,
  title={Reliability, validity, and factor structure of the creative achievement questionnaire},
  author={Carson, Shelley H and Peterson, Jordan B and Higgins, Daniel M},
  journal={Creativity research journal},
  volume={17},
  number={1},
  pages={37--50},
  year={2005}
}

@article{jauk2014road,
  title={The road to creative achievement: A latent variable model of ability and personality predictors},
  author={Jauk, Emanuel and Benedek, Mathias and Neubauer, Aljoscha C},
  journal={European journal of personality},
  volume={28},
  number={1},
  pages={95--105},
  year={2014}
}

@article{podsakoff2003common,
  title={Common method biases in behavioral research: a critical review of the literature and recommended remedies},
  author={Podsakoff, Philip M and MacKenzie, Scott B and Lee, Jeong-Yeon and Podsakoff, Nathan P},
  journal={Journal of applied psychology},
  volume={88},
  number={5},
  pages={879},
  year={2003}
}

@article{park2016revisiting,
  title={Revisiting individual creativity assessment: Triangulation in subjective and objective assessment methods},
  author={Park, Namgyoo K and Chun, Monica Youngshin and Lee, Jinju},
  journal={Creativity Research Journal},
  volume={28},
  number={1},
  pages={1--10},
  year={2016}
}

@article{zeng2011can,
  title={Can traditional divergent thinking tests be trusted in measuring and predicting real-world creativity?},
  author={Zeng, Liang and Proctor, Robert W and Salvendy, Gavriel},
  journal={Creativity research journal},
  volume={23},
  number={1},
  pages={24--37},
  year={2011}
}

@article{yang2022creative,
  title={Creative problem solving in knowledge-rich contexts},
  author={Yang, Wenjing and Green, Adam E and Chen, Qunlin and Kenett, Yoed N and Sun, Jiangzhou and Wei, Dongtao and Qiu, Jiang},
  journal={Trends in Cognitive Sciences},
  volume={26},
  number={10},
  pages={849--859},
  year={2022}
}

@article{Project_Debater,
  author	= "Noam Slonim and
                  Yonatan Bilu and
                  Carlos Alzate and
                  Roy Bar{-}Haim and
                  Ben Bogin and
                  Francesca Bonin and
                  Leshem Choshen and
                  Edo Cohen{-}Karlik and
                  Lena Dankin and
                  Lilach Edelstein and
                  Liat Ein{-}Dor and
                  Roni Friedman{-}Melamed and
                  Assaf Gavron and
                  Ariel Gera and
                  Martin Gleize and
                  Shai Gretz and
                  Dan Gutfreund and
                  Alon Halfon and
                  Daniel Hershcovich and
                  Ron Hoory and
                  Yufang Hou and
                  Shay Hummel and
                  Michal Jacovi and
                  Charles Jochim and
                  Yoav Kantor and
                  Yoav Katz and
                  David Konopnicki and
                  Zvi Kons and
                  Lili Kotlerman and
                  Dalia Krieger and
                  Dan Lahav and
                  Tamar Lavee and
                  Ran Levy and
                  Naftali Liberman and
                  Yosi Mass and
                  Amir Menczel and
                  Shachar Mirkin and
                  Guy Moshkowich and
                  Shila Ofek{-}Koifman and
                  Matan Orbach and
                  Ella Rabinovich and
                  Ruty Rinott and
                  Slava Shechtman and
                  Dafna Sheinwald and
                  Eyal Shnarch and
                  Ilya Shnayderman and
                  Aya Soffer and
                  Artem Spector and
                  Benjamin Sznajder and
                  Assaf Toledo and
                  Orith Toledo{-}Ronen and
                  Elad Venezian and
                  Ranit Aharonov",
  title	= "An autonomous debating system",
  journal	= "Nature",
  volume	= "591",
  number	= "7850",
  pages	= "379--384",
  year	= "2021"
}

@inproceedings{PersuasionForGood,
  author	= "Xuewei Wang and
                  Weiyan Shi and
                  Richard Kim and
                  Yoojung Oh and
                  Sijia Yang and
                  Jingwen Zhang and
                  Zhou Yu",
  title	= "Persuasion for Good: Towards a Personalized Persuasive Dialogue System for Social Good",
  volume	= "1",
  booktitle = "Proceedings of the 57th Conference of the Association for Computational Linguistics",
  pages = "5635--5649",
  address	= "Florence, Italy",
  year = "2019"
}

@article{Reference-Based,
  author       = {Ruizhe Li and
                  Chiwei Zhu and
                  Benfeng Xu and
                  Xiaorui Wang and
                  Zhendong Mao},
  title        = {Automated Creativity Evaluation for Large Language Models: {A} Reference-Based Approach},
  journal      = {Computing Research Repository},
  volume       = {abs/2504.15784},
  year         = {2025},
}

@article{SMOTE,
  author	= "Weng, Weini and Liu, Chang and Zhao, Guoli and Song, Luwei and Zhang, Xingli",
  title	= "Intelligent Assessment of Scientific Creativity by Integrating Data Augmentation and Pseudo-Labeling",
  journal		= "Information",
  volume		= "16",
  number		= "9",
  pages		= "785",
  year		= "2025"
}

@article{alphacontext,
  title={AlphaContext: An Evolutionary Tree-based Psychometric Context Generator for Creativity Assessment},
  author={Wang, Yixuan and Huang, Yue and Qian, Hong and Wei, Yunzhao and Ding, Yifei and Wang, Wenkai and Liu, Zhi and Huang, Zhongjing and Zhou, Aimin and Guo, Jiajun},
  journal={arXiv preprint arXiv:2604.18398},
  year={2026}
}

@inproceedings{OpenDebateEvidence,
  author	= "Allen Roush and Yusuf Shabazz and Arvind Balaji and Peter Zhang and Stefano Mezza and Markus Zhang and Sanjay Basu and Sriram Vishwanath and Mehdi Fatemi and Ravid Shwartz-Ziv",
  title		= "OpenDebateEvidence: A Massive-Scale Argument Mining and Summarization Dataset",
  booktitle	= "Advances in Neural Information Processing Systems 38",
  address		= "Vancouver,Canada",
  year			= "2024"
}

@inproceedings{IAM,
  author	= "Cheng, Liying  and
             Bing, Lidong  and
             He, Ruidan  and
             Yu, Qian  and
             Zhang, Yan  and
             Si, Luo",
  title	= "{IAM:} {A} Comprehensive and Large-Scale Dataset for Integrated Argument Mining Tasks",
  booktitle	= "Proceedings of the 60th Annual Meeting of the Association for Computational Linguistics",
  pages		= "2277--2287",
  address		= "Dublin, Ireland",
  year		= "2022"
}

@article{glassner2007stands,
  author    = "Glassner, Amnon and Schwarz, Baruch B.",
  title     = "What stands and develops between creative and critical thinking? Argumentation?",
  journal   = "Thinking Skills and Creativity",
  volume    = "2",
  year      = "2007"
}

@article{zhang2026creative,
  title="Research on Automatic Evaluation of Idea Quality in Knowledge Building Communities",
  author="Yiwen Zhang and Hong Qian and Xiaowen Wang and Yixvan Wang and Mingjia Li and Jin Wu and Jiajun Guo and Xiangfeng Wang and Chanjin Zheng and Aimin Zhou",
  journal="China Educational Technology",
  number="472",
  pages="85--94",
  year="2026"
}

@inproceedings{LLM-as-a-Judge,
  author		= "Lianmin Zheng and Wei-Lin Chiang and Ying Sheng and Siyuan Zhuang and Zhanghao Wu and Yonghao Zhuang and Zi Lin and Zhuohan Li and Dacheng Li and Eric P. Xing and Hao Zhang and Joseph E. Gonzalez and Ion Stoica",
  title		= "Judging LLM-as-a-Judge with MT-Bench and Chatbot Arena",
  booktitle	= "Advances in Neural Information Processing Systems 36",
  address		= "New Orleans, LA",
  year			= "2023"
}

@article{Dialectical_Agent,
  author		= "Anghel, Catalin and Anghel, Andreea Alexandra and Pecheanu, Emilia and Susnea, Ioan and Cocu, Adina and Istrate, Adrian",
  title		= "Multi-Model Dialectical Evaluation of LLM Reasoning Chains: A Structured Framework with Dual Scoring Agents",
  journal		= "Informatics",
  volume		= "12",
  pages			= "76",
  year			= "2025"
}

@inproceedings{Empirical_Analysis,
  author	= "Xinyi Liu and
                  Pinxin Liu and
                  Hangfeng He",
  title	= "An Empirical Analysis on Large Language Models in Debate Evaluation",
  booktitle	= "Proceedings of the 62nd Annual Meeting of the Association for Computational Linguistics",
  pages		= "470--487",
  address		= "Bangkok, Thailand",
  year		= "2024"
}

@article{Robust,
  author	= "Song, Hwanjun and Kim, Minseok and Park, Dongmin and Shin, Yooju and Lee, Jae-Gil",
  title	= "Learning From Noisy Labels With Deep Neural Networks: A Survey",
  journal	= "IEEE Transactions on Neural Networks and Learning Systems",
  volume	= "34",
  pages	= "8135-8153",
  year	= "2023"
}

@inproceedings{ORCHID,
  author = {Mahdi Karami and Ali Ghodsi},
  title = {Orchid: Flexible and Data-Dependent Convolution for Sequence Modeling},
  booktitle = {Advances in Neural Information Processing Systems 38},
  address = {Vancouver,Canada},
  year = {2024}
}

@inproceedings{ClaimRev,
  author		= "Gabriella Skitalinskaya and
                  Jonas Klaff and
                  Henning Wachsmuth",
  title		= "Learning From Revisions: Quality Assessment of Claims in Argumentation at Scale",
  booktitle	= "Proceedings of the 16th Conference of the European Chapter of the Association for Computational Linguistics",
  pages		= "1718--1729",
  address		= "Virtual",
  year			= "2021"
}

@article{Thin_slices_of_creativity,
  title={Thin slices of creativity: Using single-word utterances to assess creative cognition},
  author={Prabhakaran, Ranjani and Green, Adam E and Gray, Jeremy R},
  journal={Behavior research methods},
  volume={46},
  number={3},
  pages={641--659},
  year={2014}
}

@article{Semantic_distance,
  title={Semantic distance: An automated measure of creativity that is novel and appropriate},
  author={Heinen, David JP and Johnson, Dan R},
  journal={Psychology of Aesthetics, Creativity, and the Arts},
  volume={12},
  number={2},
  pages={144},
  year={2018}
}

@article{assessment_with_SemDis,
  title={Automating creativity assessment with SemDis: An open platform for computing semantic distance},
  author={Beaty, Roger E and Johnson, Dan R},
  journal={Behavior research methods},
  volume={53},
  number={2},
  pages={757--780},
  year={2021}
}

@article{application_by_elaboration,
  title={Application of latent semantic analysis to divergent thinking is biased by elaboration},
  author={Forthmann, Boris and Oyebade, Oluwatosin and Ojo, Adebusola and G{\"u}nther, Fritz and Holling, Heinz},
  journal={The Journal of Creative Behavior},
  volume={53},
  number={4},
  pages={559--575},
  year={2019}
}

@article{Automated_scoring,
  title={Beyond semantic distance: Automated scoring of divergent thinking greatly improves with large language models},
  author={Organisciak, Peter and Acar, Selcuk and Dumas, Denis and Berthiaume, Kelly},
  journal={Thinking Skills and Creativity},
  volume={49},
  pages={101356},
  year={2023}
}

@article{scoring_with_LLM,
  title={Automatic scoring of metaphor creativity with large language models},
  author={DiStefano, Paul V and Patterson, John D and Beaty, Roger E},
  journal={Creativity Research Journal},
  volume={37},
  number={4},
  pages={555--569},
  year={2025}
}

@article{luchini2025automated,
  title={Automated scoring of creative problem solving with large language models: A comparison of originality and quality ratings},
  author={Luchini, Simone A and Maliakkal, Nadine T and DiStefano, Paul V and Laverghetta Jr, Antonio and Patterson, John D and Beaty, Roger E and Reiter-Palmon, Roni},
  journal={Psychology of Aesthetics, Creativity, and the Arts},
  year={2025}
}

@inproceedings{15-Dimensional,
  author		= "Henning Wachsmuth and
                  Nona Naderi and
                  Ivan Habernal and
                  Yufang Hou and
                  Graeme Hirst and
                  Iryna Gurevych and
                  Benno Stein",
  title		= "Argumentation Quality Assessment: Theory vs. Practice",
  volume		= "2",
  booktitle		= "Proceedings of the 55th Annual Meeting of the Association for Computational Linguistics",
  pages			= "250--255",
  address		= "Vancouver, Canada",
  year			= "2017"
}

@article{creative_processes,
  title={Creative thinking processes: The past and the future},
  author={Mumford, Michael D and McIntosh, Tristan},
  journal={The Journal of Creative Behavior},
  volume={51},
  number={4},
  pages={317--322},
  year={2017},
  publisher={Wiley Online Library}
}

@article{process_analytic_models,
  title={Process analytic models of creative capacities},
  author={Mumford, Michael D and Mobley, Michele I and Reiter-Palmon, Roni and Uhlman, Charles E and Doares, Lesli M},
  journal={Creativity research journal},
  volume={4},
  number={2},
  pages={91--122},
  year={1991}
}

@article{intellect_structure,
  title={The structure of intellect.},
  author={Guilford, Joy Paul},
  journal={Psychological bulletin},
  volume={53},
  number={4},
  pages={267},
  year={1956}
}

@article{torrance,
  title={Torrance tests of creative thinking},
  author={Torrance, E Paul},
  journal={Educational and psychological measurement},
  year={1966}
}

@article{creativity_standard,
  title={The standard definition of creativity},
  author={Runco, Mark A and Jaeger, Garrett J},
  journal={Creativity research journal},
  volume={24},
  number={1},
  pages={92--96},
  year={2012},
}

@article{intelligence_nature,
  title={The Nature of Human Intelligence},
  author={ Guilford, J. Paul },
  journal={McGraw-Hill,},
  year={1967},
}

@article{convergent_thinking,
  title={In praise of convergent thinking},
  author={Cropley, Arthur},
  journal={Creativity research journal},
  volume={18},
  number={3},
  pages={391--404},
  year={2006}
}

@article{appraisal_revision_standards,
  title={Evaluative aspects of creative thought: Effects of appraisal and revision standards},
  author={Lonergan, Devin C and Scott, Ginamarie M and Mumford, Michael D},
  journal={Creativity Research Journal},
  volume={16},
  number={2-3},
  pages={231--246},
  year={2004}
}

@inproceedings{Construction_Analysis,
  author       = {Shai Gretz and
                  Roni Friedman and
                  Edo Cohen{-}Karlik and
                  Assaf Toledo and
                  Dan Lahav and
                  Ranit Aharonov and
                  Noam Slonim},
  title = {A Large-Scale Dataset for Argument Quality Ranking: Construction and Analysis},
  booktitle = {Proceedings of the 34th {AAAI} Conference on Artificial Intelligence},
  pages   = {7805--7813},
  address = {New York, NY},
  year   = {2020}
}

@article{predictive_model_of_debate,
  author       = {Lu Wang and
                  Nick Beauchamp and
                  Sarah Shugars and
                  Kechen Qin},
  title        = {Winning on the Merits: The Joint Effects of Content and Style on Debate Outcomes},
  journal={Transactions of the Association for Computational Linguistics},
  volume       = {5},
  pages        = {219--232},
  year         = {2017}
}

@article{InspireDebate,
  title={InspireDebate: Multi-Dimensional Subjective-Objective Evaluation-Guided Reasoning and Optimization for Debating},
  author={Wang, Fuyu and Li, Jiangtong and Zhu, Kun and Jiang, Changjun},
  journal={arXiv preprint arXiv:2506.18102},
  year={2025}
}

@inproceedings{Debatrix,
  title={Debatrix: Multi-dimensional debate judge with iterative chronological analysis based on llm},
  author={Liang, Jingcong and Ye, Rong and Han, Meng and Lai, Ruofei and Zhang, Xinyu and Huang, Xuan-Jing and Wei, Zhongyu},
  booktitle={Findings of the 62nd Annual Meeting of the Association for Computational Linguistics},
  pages={14575--14595},
  address	= {Bangkok, Thailand},
  year={2024}
}

@inproceedings{LoRA,
  author       = {Edward J. Hu and
                  Yelong Shen and
                  Phillip Wallis and
                  Zeyuan Allen{-}Zhu and
                  Yuanzhi Li and
                  Shean Wang and
                  Lu Wang and
                  Weizhu Chen},
  title        = {LoRA: Low-Rank Adaptation of Large Language Models},
  booktitle    = {Proceedings of the 10th International Conference on Learning Representations},
  address   = {Virtual},
  year         = {2022},
}

@inproceedings{themis,
  title={Themis: A reference-free nlg evaluation language model with flexibility and interpretability},
  author={Hu, Xinyu and Lin, Li and Gao, Mingqi and Yin, Xunjian and Wan, Xiaojun},
  booktitle={Proceedings of the 2024 Conference on Empirical Methods in Natural Language Processing},
  pages={15924--15951},
  year={2024}
}

@article{mprometheus,
  title={M-Prometheus: A Suite of Open Multilingual LLM Judges},
  author={Pombal, Jos{\'e} and Yoon, Dongkeun and Fernandes, Patrick and Wu, Ian and Kim, Seungone and Rei, Ricardo and Neubig, Graham and Martins, Andr{\'e} FT},
  journal={arXiv preprint arXiv:2504.04953},
  year={2025}
}

@article{creval,
  title={Evaluating text creativity across diverse domains: A dataset and large language model evaluator},
  author={Cao, Qian and Wang, Xiting and Yuan, Yuzhuo and Liu, Yahui and Luo, Fang and Song, Ruihua},
  journal={arXiv preprint arXiv:2505.19236},
  year={2025}
}

@article{Qwen2.5,
  title={Qwen2.5-vl technical report},
  author={Bai, Shuai and Chen, Keqin and Liu, Xuejing and Wang, Jialin and Ge, Wenbin and Song, Sibo and Dang, Kai and Wang, Peng and Wang, Shijie and Tang, Jun and others},
  journal={arXiv preprint arXiv:2502.13923},
  year={2025}
}

\appendix
\section*{Appendix}
\begin{figure*}[ht]
\centering
\includegraphics[width=0.62\textwidth]{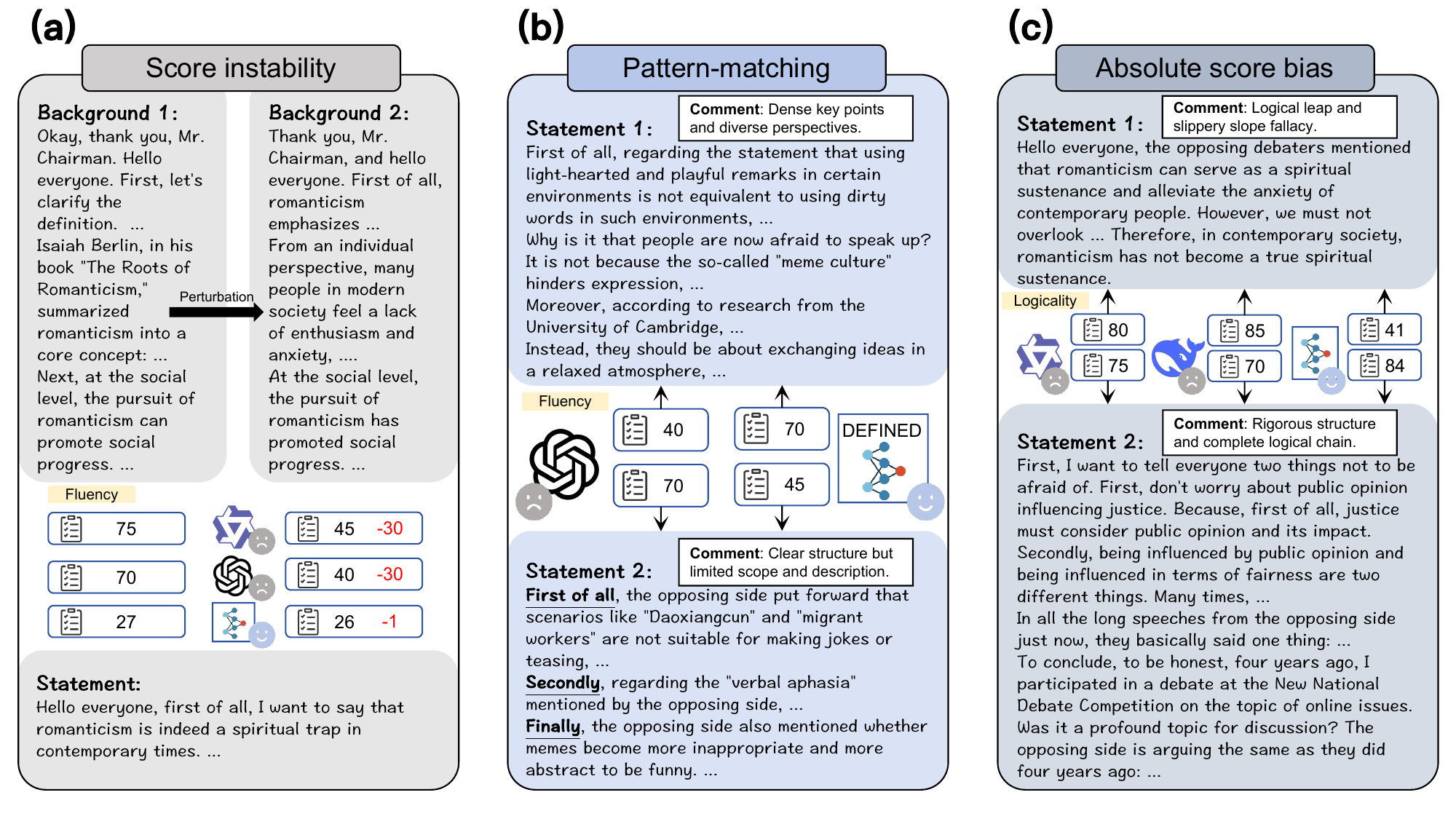}
\caption{\textbf{Case study of typical bias patterns in LLM-based scoring and the robustness of DEFINED.} \textbf{(a)} Score instability under contextual perturbation. \textbf{(b)} Surface-level pattern matching. \textbf{(c)} Absolute score mismatch.}
\Description{pic}\label{case_study_pic}
\end{figure*}
\section{Detailed Metric System}\label{definition}
Detailed definition of eight dimensional metric system: 
\begin{itemize}

   \item \textbf{Originality (Creativity).}  
  Originality refers to the relative uniqueness or rarity of ideas or solutions proposed within a given context~\cite{intellect_structure,torrance,creativity_standard}. In debates, this dimension concerns the distinctiveness of claims, analogies, or argumentative pathways. It requires extending or transforming foundational positions when deconstructing opponents’ arguments or articulating one’s own stance~\cite{Dialectical_Agent}.

  \item \textbf{Fluency (Creativity).}  
  Fluency is defined as the capacity to rapidly and continuously generate a large number of distinct ideas, responses, or potential solutions in open-ended settings~\cite{intellect_structure,torrance}. In debate scenarios, it is operationalised as the number of relevant arguments or rebuttal points produced under given time and information constraints.

  \item \textbf{Flexibility (Creativity).} 
  Flexibility refers to the ability to shift across categories, perspectives, or reasoning pathways during the thinking process, generating diverse types of answers or solutions~\cite{intelligence_nature,torrance}. In debates, this dimension reflects cross-domain or multi-strategy reasoning, such as approaching an issue from legal and economic perspectives, moving between macro- and micro-level analyses.

    \item \textbf{Relevance (Creativity).}  
  In creativity research, appropriateness denotes the degree to which an idea fits the task context, goals, and constraints ~\cite{creativity_standard}. Within this broader construct, relevance denotes the ability to focus on the core issues within complex problems and to accurately identify unresolved or salient points that warrant further examination. In debates, it refers to identifying critical weaknesses in the opponent’s reasoning or direct challenges to one’s own position, and responding explicitly to the central points of contention without digression~\cite{15-Dimensional}.

  \item \textbf{Logicality (Creativity).}  
  In cognitive processes, reasoning follows coherent inferential pathways, applying prior knowledge and rules to analyse information and derive judgments or conclusions in a logically consistent manner~\cite{convergent_thinking,appraisal_revision_standards}. In a debate context, logicality refers to the ability to construct a clear and complete chain of reasoning, organise arguments in a structured way, avoid logical fallacies or inappropriate evidence, and explicitly articulate the relationships between claims and supporting evidence~\cite{ClaimRev, Construction_Analysis}.
  
  \item \textbf{Clarity.}  
  Clarity denotes the degree to which information is presented in an intelligible and accessible manner, avoiding ambiguity or vagueness, thereby enabling the audience to quickly and accurately grasp the core message~\cite{Dialectical_Agent,15-Dimensional}.

  \item \textbf{Appeal.}  
  Appeal captures the overall impact of an argument on the audience at both the content and stylistic levels. It reflects the strategic use of emotional expression or contextualised narrative to foster resonance with listeners and thereby enhance persuasive force~\cite{predictive_model_of_debate, 15-Dimensional}.

  \item \textbf{Effectiveness.}  
  Effectiveness measures whether conclusions or strategic choices, while remaining logically consistent and task-relevant, successfully address the issue at hand and advance the argumentative process, ultimately achieving the practical goal of refuting the opponent or reinforcing one’s own position~\cite{15-Dimensional}.
\end{itemize}

\begin{figure*}[ht]
\centering
\includegraphics[width=0.61\textwidth]{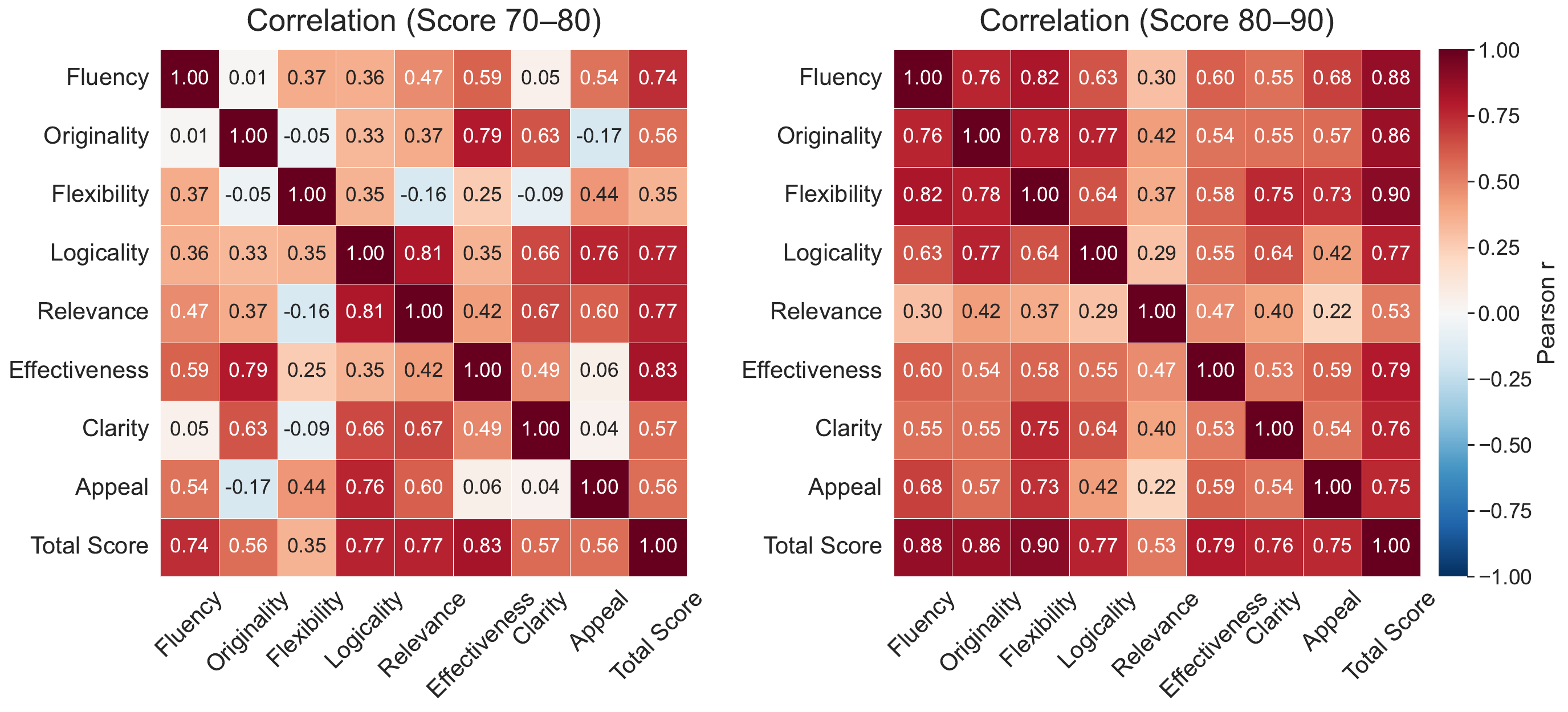}
\caption{\textbf{Inter-dimension correlation matrix of eight-dimensional metric system.}}
\Description{pic}\label{corr_appendix}
\end{figure*}

\section{Augmentation Prompt}\label{Augmentation_prompt}
Due to space limitations, we present one dimension prompt as an example. Other dimensions follow the same prompt structure.
\begin{description}[style=nextline, leftmargin=1.0em, font=\bfseries]

\item[Level 40 (Basic Capability)]
\textit{Instruction A:} You have basic argumentation skills. Please strictly follow these requirements: You can only propose 1 relevant point revolving around the topic or the opponent's view, but with limited depth. Your viewpoint is common, lacking the ability to innovate or redefine the problem. You can maintain roughly coherent logic, but often with jumps or parts lacking explanation. You tend to focus on a single domain or a single type of reasoning path and do not actively switch analytical angles. When responding to the opponent, you might notice a simple loophole but cannot explain its importance or build an effective counterattack. Content is dominated by shallow analysis, completely failing to build systematic frameworks or multiple logical chains, maintaining only rough discourse.

\textit{Instruction B:} You have basic but weak argumentation skills. You are allowed to propose around 1 relevant point (at most 1 main argument + some qualifying explanations), without a second independent argument. The main argument should be a common, obvious conclusion, expressed in common frameworks (causality or comparison); explanations must be limited and superficial, without digging deep or building complete reasoning chains. You may notice a simple loophole in the opponent's argument but can only point out its existence, not explain its mechanism or scope of impacts. Prohibit cross-domain switching, creating new concepts, or complex analogies; do not use precise data or complex evidence. Output should be roughly coherent but may have jumps or lack explanation. Strict limit: No more than 1 independent main argument; response to the opponent is limited to ``mentioning'' rather than ``deep refutation''.

\end{description}

\section{Case Study}\label{case_study}
Detailed demonstration of case studies on typical bias patterns in the scoring behavior of large language models is shown in Fig~\ref{case_study_pic}.

\section{Fine-grained Prompt}\label{fine_prompt}

Due to space limitations, we present only one dimension prompt as an example. The remaining dimensions follow the same prompt structure, including theoretical definition, operational definition, scoring rubrics, checkpoints, and illustrative examples.

\textbf{Dimension: Divergent Thinking --- Originality}

\textbf{Theoretical Definition.} 
The relative uniqueness or rarity of the ideas or solutions proposed within a given context, distinguishing them from conventional or obvious options.

\textbf{Operational Definition.} 
The uniqueness of the discourse, analogies, or argumentative paths. It requires extending or transforming the foundational arguments rather than simply restating conventional logic.

\textbf{Scoring Rubrics.}
\begin{itemize}
    \item \textbf{90--100:} Presents a distinctly unique reasoning path; proposes rare but reasonable new perspectives or creative frameworks that redefine the issue.
    
    \item \textbf{70--89:} Offers some novelty in perspective; extends or refines common views, though the innovation is limited to specific segments rather than the overall framework.
    
    \item \textbf{40--69:} Largely follows common argumentative paths with occasional minor variations; lacks systematic innovation.
    
    \item \textbf{0--39:} Primarily restates the prompt or repeats existing positions without new insights; relies on ready-made logic.
\end{itemize}
\textbf{Checkpoints (Scoring Reference).}

\begin{enumerate}
    \item Presence of unique or rare argumentative angles?
    
    \item Successful extension or transformation of the original argumentative base? 
    
    \item Integration of new analogies, concepts, or reasoning chains?
\end{enumerate}

\textbf{Example.}

\begin{itemize}
    \item \textbf{High Score:} Reinterprets AI not as a competition for knowledge, but as a ``power shift in defining intellectual authority'', thereby constructing an original structural argument.
    
    \item \textbf{Low Score:} Merely repeats common clichés (e.g., ``AI lacks emotion'') without introducing new perspectives or argumentative transformations.
\end{itemize}

\section{Inter-dimension Correlation Matrix}\label{matrix}
To further examine the independence and discriminability of the proposed evaluation dimensions, we compute Pearson correlation matrices between all dimensions on the test set under different score intervals (Fig.~\ref{corr_appendix}). The analysis reveals that several dimensions exhibit relatively weak correlations despite belonging to the same evaluation framework. For example, within the 80--90 score range, the correlation between Relevance and Logicality is only 0.15, while in the 70--80 range, the correlation between Effectiveness and Appeal decreases to 0.06. These results indicate that the proposed dimensions capture distinct cognitive aspects of debate creativity rather than redundant information.

\end{document}